\documentclass[a4paper, conference]{IEEEtran}
\IEEEoverridecommandlockouts

\title{
A Concept for User-Centered Delegation of Abstract High-Level Tasks to Cobots for Flexible Lot Sizes\\
\thanks{This publication was created as part of the AI Production Network
Augsburg research project at the Augsburg Technical University of Applied Sciences.}
}
\author{\IEEEauthorblockN{Moritz Schmidt \href{https://orcid.org/0009-0008-5666-9666}{\includegraphics[height=8pt]{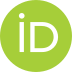}}}
\IEEEauthorblockA{\textit{Faculty of Electrical Engineering} \\
\textit{Augsburg Technical University of Applied Sciences}\\
Augsburg, Germany \\
moritz.schmidt@hs-augsburg.de}
\and
\IEEEauthorblockN{Claudia Meitinger}
\IEEEauthorblockA{\textit{Faculty of Electrical Engineering} \\
\textit{Augsburg Technical University of Applied Sciences }\\
Augsburg, Germany \\
claudia.meitinger@hs-augsburg.de}
}

\usepackage{cite}
\usepackage{amsmath,amssymb,amsfonts}
\usepackage{algorithmic}
\usepackage{graphicx}
\usepackage{textcomp}
\usepackage{xcolor}
\usepackage[T1]{fontenc}

\usepackage[hyphens]{url}
\usepackage[hidelinks]{hyperref}
\usepackage{cleveref}
\usepackage{graphicx}
\usepackage{multicol}
\usepackage[acronym]{glossaries}

\def\BibTeX{{\rm B\kern-.05em{\sc i\kern-.025em b}\kern-.08em
    T\kern-.1667em\lower.7ex\hbox{E}\kern-.125emX}}

\makeglossaries
\newacronym{sme}{SME}{Small and Medium Enterprise}
\newacronym{ai}{AI}{Artificial Intelligence}
\newacronym{ui}{UI}{User Interface}
\newacronym{ccu}{CCU}{Cognitive Control Unit}
\newacronym{pbd}{PbD}{Programming by Demonstration}
\newacronym{hmi}{HMI}{Human Machine Interface}
\newacronym{uri}{URI}{Unique Resource Identifier}
\newacronym{owl}{OWL}{Web Ontology Language}

\usepackage{todonotes}

\begin{document}

\maketitle

\begin{abstract}
   Technical advances in collaborative robots (cobots) are making them
   increasingly attractive to companies. However, many human operators are not
   trained to program complex machines. Instead, humans are used to 
   communicating with each other on a task-based level rather than through
   specific instructions, as is common with machines. The gap between low-level
   instruction-based and high-level task-based communication leads to low
   values for usability scores of teach pendant programming. As a
   solution, we propose a task-based interaction concept that allows human
   operators to delegate a complex task to a machine without programming by
   specifying a task via triplets. The concept is based on task decomposition
   and a reasoning system using a cognitive architecture. The approach is
   evaluated in an industrial use case where mineral cast basins have to be
   sanded by a cobot in a crafts enterprise.\\
\end{abstract}

\begin{IEEEkeywords}
    Human-robot interaction, Task-based interaction, human-centered automation
\end{IEEEkeywords}


\section{Statement of the Problem}

Collaborative robots are becoming more and more attractive for \glspl{sme} due to
ongoing technological progress, but it is often not yet possible to use cobots
efficiently and effectively in the area of single-unit and small-lot
production due to a gap in the level of abstraction with respect to communication 
between the human operator and cobot.

The configuration of a cobot to handle a new task often requires programming or
parameterization of predefined skills. Human operators, usually shop floor
workers, are often not trained in programming whereby a gap in competencies is
created. This limits potential applications in the field, as either the
human operator would have to be trained in a new skill - programming - or the
communication between human operators and cobots have to change.

For this reason, we propose an abstract task-based interaction paradigm
transferring the programming workload from human operators to automated
programs and algorithms. This approach does not require the human operator to
program a machine online on a primitive instruction or skill basis. Instead,
abstract tasks based on triplets, e.g. \texttt{sand - mineral cast - basin} are
provided by a \gls{ui} and decomposed by a \gls{ccu} \cite{Faber2017} into
sub-tasks. Separated tasks are executed on the machine via a task-oriented
programming approach. As a consequence, a human operator is no longer required
to program or parameterize machines online.

This paper is structured as follows: first, the research context is presented,
cf. \cref{sec:context}. Afterwards, related use cases are depicted in
\cref{sec:use-case}, followed by related work, cf.
\cref{sec:state-of-the-art}. Based on the related work, a concept for
high-level task-based interaction is presented in \cref{sec:concept}, and contributions,
cf. \cref{sec:contributions}, are described. 

\section{Challenges of Small and Medium Crafts Enterprises}
\label{sec:context}

\IEEEpubidadjcol 

Our research on task-based interaction with cobots focuses on \gls{sme} for a
number of reasons: (1) increasing skilled labor shortage, (2) \gls{ui}s with
low usability scores, (3) single-unit production, and (4) a strong \gls{sme}
industry in Germany.

German labor market data shows an increasing challenge in filling vacancies due
to demographic developments, among other things \cite{Arbeit2022, Arbeit2022a}.
The retirement rate is going to reach its maximum between 2023 and 2027
further increasing the lack of skilled workers \cite{Forster2019}. Although
this data is only applicable to the demographic development in Germany, similar
developments can be expected in other countries with an aging population
\cite{Rosling2021}.

Giannopoulou et al. found in a user study that human operators are not keen on
programming cobots or would do it only if they are required to do so. A
minority stated they would enjoy getting into programming cobots
\cite{Giannopoulou2021}. This is also related to the user-study of Dong et al.
where cobot teaching by means of a teach pendant resulted in a system usability  
score below average \cite{Dong2021}.
The findings of Giannopoulou et al. and Dong et al. rely on user studies with small sample
sizes, therefore they must be considered carefully due to the findings of a meta
study conducted by Leichtmann et al. The results suggest that most user studies
are prone to "small sample sizes, lack of theory or missing information in
reported data" \cite{Leichtmann2022} and cannot be replicated \cite{Leichtmann2022}.

Single-unit production is prevalent in \glspl{sme} due to a variety of factors. One
key reason is that \glspl{sme} often operate in niche markets or produce specialized
products, which are not well-suited for large lot sizes. 

Furthermore, \gls{sme} account for 37.2 \% of the summarized turnover of all
companies in Germany, with 58.3 \% of all employees working for \glspl{sme}
\cite{Forster2019}. Zimmermann et al. reported that micro and small
enterprises are increasingly investing less in digitization and could therefore
be left behind by medium-sized enterprises with increasing spending on
digitization and additionally on research and development \cite{Zimmermann2022a}.

\section{Use Cases}
\label{sec:use-case}

The selected use cases are based on previously conducted observations
(shadowings) in crafts enterprises. Firstly, a carpenter's workshop, producing
a flexible lot size of mineral cast basins  each year and secondly, a
company constructing conservatories, screwing aluminum connectors to beams in order to
connect the basic structure.

\subsection{Sanding Mineral Cast Basins}
\label{sec:use-case_sanding}

The manufacturing process of mineral cast involves a lot of repetitive, dusty
and noisy sanding of partially complex shapes. In particular, the surfaces must
be sanded up to seven times with sanding paper of varying granularity to reach
the desired surface quality. The sanding task is
often delegated to workers with low salaries, whereas it is a use case for a
cobot application - long-lasting, monotonous, loud and dusty work.

While the cobot sands the mineral cast basin, the craftsman could pursue other
tasks and return to the cobot once the task is completed. In case the mineral
cast basin does not pass the surface quality check by a human operator, the
cobot could be instructed to do minor reworking for specific regions.

\subsection{Screwing Aluminum Connectors for Conservatories}

Constructing conservatories requires connecting several beams of wood for
building the structure responsible for holding the roof as well as glass walls.
The beams are connected by dovetail aluminum connectors screwed to the beams in
designated positions which are already milled in a previous production step.
Each connector requires four screws for alignment and then up to fourteen
screws to ensure a sufficient force transmission. As a result, a single
conservatory can consist of several hundred or even up to ten thousand screws. 

A craftsman is required to tighten the screws for alignment whereas the
remaining screws make up a monotonous and long-lasting task which could be
conducted by a cobot trailing the craftsman.

\section{Related Work}
\label{sec:state-of-the-art} 

The use cases could already be implemented today, but the level of
communication where craftsmen communicate among themselves is more abstract
compared with the programming approaches used to instruct cobots. At the
moment, industry solutions for cobot programming are based on primitive
instructions or simple skill-based structures. Therefore, other cobot
programming approaches must be pursued in order to bridge the gap between the
level of communication between cobot and craftsmen. The need for efficient
programming approaches becomes particularly noticeable when single-unit
productions should be handled with a cobot, because the setup and programming
time might exceed the time of manually completing the task at hand.

\subsection{Human-Cobot Interfaces and Programming Approaches} 
\label{sec:hcpi}

In the past several approaches have been implemented to simplify the
programming of cobots via different programming approaches, e.g. \gls{pbd},
task-oriented programming or task-based programming. A key feature of
task-oriented programming is the automatic generation of cobot programs based
on a task description \cite{Siegert1996, Humburger1998, Weber2022}, whereas
task-based programming could rely on offline-defined programs or skills which
require online parameterization, for instance \cite{Steinmetz2016}.

Schou et al. presented an approach based on three levels of abstractions,
device primitives, skills and tasks \cite{Schou2013}, inspired by \cite{Gat1998}. Human
operators can program the robot via an \gls{ui} in two steps: task specification and
\gls{pbd} for final parameterization \cite{Schou2013}. Furthermore, Schou et
al. devised a skill model representing the effects, pre- and postconditions of skills on
the environment \cite{Schou2013, Schou2018}. A similar approach is followed by
\cite{Pedersen2014, Steinmetz2016, Steinmetz2018}. Additionally, Koch et al. extended
the \textit{task - skill - primitive instruction} abstraction hierarchy from
Schou et al. by a more abstract mission level \cite{Koch2017}. 

Steinmetz et al. diverge from the previous approach by following a semantic
skill recognition procedure during a \gls{pbd} approach in order to
automatically create a task sequence in hindsight \cite{Steinmetz2019}.
Guerin et al. proposed a \gls{ui} for high-level user functions and \gls{pbd}
approaches, playback functionality and visualizing constraints \cite{Guerin2014}.
Paxton et al. followed an approach based on behavior trees and multiple
corresponding \gls{ui}s where only detected objects are displayed and
corresponding moves, waypoints, and configuration of additional moves can be
parameterized \cite{Guerin2015, Paxton2017}. 

Cobot teaching methods usually require programming or online parameterization
via a \gls{hmi} and might not follow a minimalistic design principle, e.g. \cite{Paxton2017},
as suggested by \cite{Nielsen1990, Norman2013, Prati2022}. 

\subsection{System Architectures for Simplified Cobot Programming Approaches}
\label{sec:sys_arch}

The underlying software architectures enable human operators to program a
cobot more efficient are usually based on a cognitive architecture, ontology, large language model or 
decision tree approach.

Zaeh et al. defined the cognitive factory to be the next step in automation for
flexible production for small lot sizes and product variations in contrast to
fully automated mass production with specified lead times \cite{Zaeh2009, Bannat2011}.
The architecture suggested by \cite{Zaeh2009, Bannat2011} resembles
the requirements for cognitive robotics defined in \cite{Shimoda2021,
Sandini2021}: perception, attention, anticipation for action and outcome,
planning, learning and adaption to change.
Faber et al. suggested an architecture based on the Soar cognitive architecture and 
an adjacent graph-based assembly sequence planner (GASP) for an assembly task \cite{Faber2017}.
Further research projects focus on a cognition-based architecture
\cite{Dang2017, Sorokoumov2021, Rovbo2022, Mininger2021}.

Perzylo et al. conducted research about full production line integration into
an \gls{owl}2 based ontology \cite{Perzylo2020}. Angleraud et al. build their
infrastructure based on an \texttt{<action, target>} pair as well as on an
\gls{owl}2 based ontology \cite{Angleraud2021} and Albu-Schäffer et al. follow
a knowledge-based engineering approach also based on an \gls{owl}2 ontology for
automatic task allocation suggestions according to "economic and human work
design criteria" \cite{AlbuSchaeffer2023}. Albu-Schäffer et al. also criticize
partial automation because of the risk that employees will become de-qualified
as more and more cognitively demanding tasks are taken over by machines
\cite{AlbuSchaeffer2023}. On the other hand, the conducted work by human
operators will change towards monitoring, deciding and organizing
\cite{Forster2019, AlbuSchaeffer2023}.

The research project of Berg et al. focuses on a fully integrated system where
the human operator is presented with a task allocation and assembly plan derived
from CAD models of the assembly and the environment for confirmation or
adoption of the generated assembly plan. The human operator is not required to
do any programming since the cobot programs are generated based on a
task-oriented programming \cite{Siegert1996, Humburger1998, Weber2022} approach
that relies on task information from an XML file \cite{Berg2017,Berg2018,
Berg2020}.

Recent research in artificial intelligence introduced large language models
which are also used in robotic applications, e.g. Palm-E architecture by Google
\cite{Driess2023}. This approach relies on large training data sets which might
not be available in the context of \gls{sme}, e.g. due to small lot sizes. In
contrast to Palm-E, Isaac Cortex by NVIDIA as a decision framework has the
potential to work with synthetic sensor data from Isaac Sim, therefore does not
have the potential lack of training data. Isaac Cortex is currently under
development \cite{Ratliff2022}. 

Related research projects already focus on simplifying the programming of
cobots, but to the best of our knowledge no approaches align with our
observations where human operators communicate via high-level task-based
commands. Nonetheless, the results of related research projects can be used 
to implement a task delegation to cobots on an abstract level. 

Koch et al. introduced the mission abstraction level, nonetheless the human
operator has to program missions on a task- and skill-based \gls{ui}
\cite{Koch2017}. Berg et al. use an offline definition of assembly sequences
and tasks on a file-based approach \cite{Berg2017} which could be utilized to
create missions for the approach proposed by Koch et al. Furthermore, the
approach by Berg et al. enables the user to decide, whether the automatically
generated sequence is acceptable. The sequence, execution, and results of skills
could be computed and monitored by a \gls{ccu} described by Faber et al., which
is yet only validated by means of a simulation study \cite{Faber2017}. Although
the \gls{ui} of Paxton et al. is very cluttered, the approach of displaying
only possible combinations of objects in the workspace could increase the user
experience because irrelevant information in the \gls{ui} can be removed
\cite{Paxton2017}.

\section{Concept}
\label{sec:concept}

\begin{figure}
    \centering
    \includegraphics[width=\linewidth,page=1]{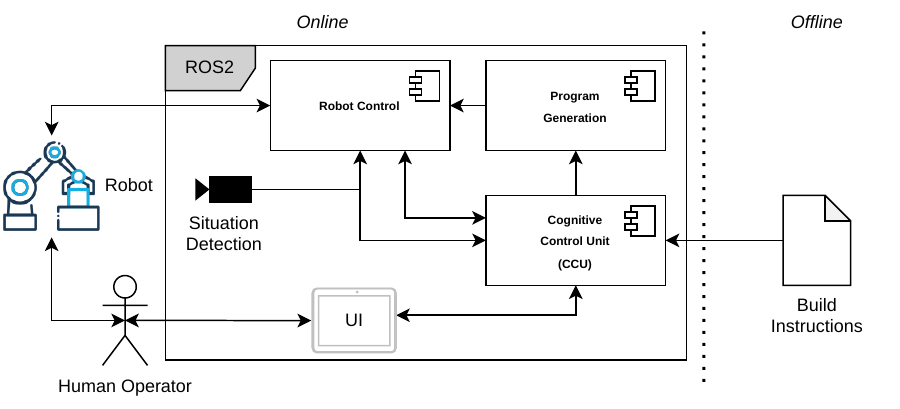}
    \caption{Component diagram}
    \label{fig:component}
\end{figure}

Our concept focuses on simplified, intuitive task delegation from human
operators to machines. \Cref{fig:component} presents our approach on a 
component-based level including robot control, program generation, situation detection,
\gls{ccu} and \gls{ui} components. 

It is based on approaches mentioned in \cref{sec:state-of-the-art}, but does
not require online parameterization or teaching. Thus, the setup time can be
reduced and possible applications for crafts enterprises can be increased.
Relying on a cognitive architecture also introduces the possibility to 
explain decisions because of the transparency of white-box approaches. Finally,
the interaction concept on the level of the \gls{ui} differs significantly in
that no sequence of capabilities has to be arranged, but only a combination of
material, process, and object has to be selected.

The robot control and program generation components are necessary but are no
focus of this research project, because automatic tool-path generation based on
point clouds can be considered state-of-the-art \cite{Zhen2019}. Besides the
fact that the \gls{ccu} will be implemented in Soar, the generation of internal
working memory elements in Soar based on build instructions will not be
implemented since this only requires parsing of documents, e.g. XML
or PDDL to Soar rules. The situation detection module is a component usually
required in a cognition-based approach but in this case, the purpose of the
situation detection is to generate a point cloud of objects contained in tasks
for the program generation module. 

The robot model is a UR10e equipped with an OnRobot Sander. 
The communication is established via ROS2 \cite{Macenski2022}.

\begin{figure*}
    \centering
    \includegraphics[width=.9\linewidth, page=2]{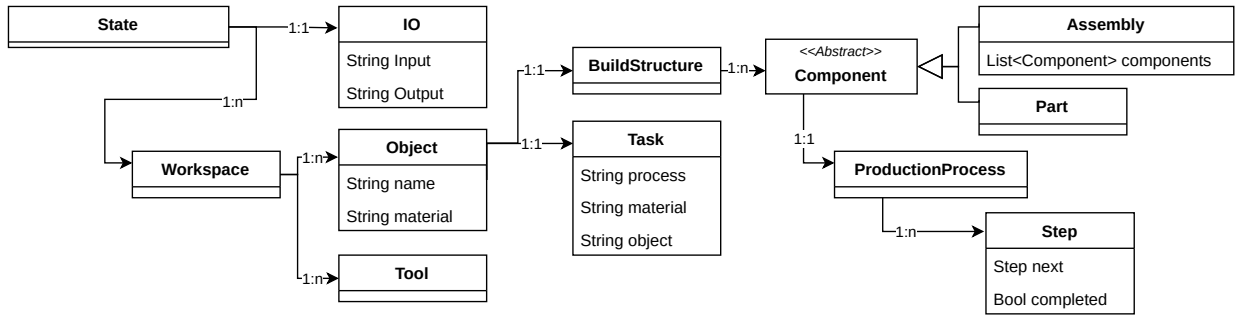}
    \caption{Proposed Soar working memory structure. The diagram is reduced to relevant data.}
    \label{fig:workingMemory}
\end{figure*}

\subsection{User Interface} 
\label{sec:ui}

The \gls{ui} design must enable users to delegate tasks to the system in an efficient,
intuitive, and understandable way. This requires several conditions to be met.
First, compliance with applicable standards and known guidelines; second, the
efficient creation of complex commands; third, the display of the system
status and running processes for the user; and fourth, the consideration of
environmental constraints.

Applicable standards for \gls{ui} design can be considered as Norman, Nielsen and
selected parts of the ISO 9241 standard. 

Preliminary user studies in crafts enterprises showed that craftsman delegate
tasks among each other on a very abstract and high level since they are aware
of each other's education and knowledge, cf. \cite{Clark2008}. In case
craftsmen have to interact with an intelligent machine, they might expect a
similar behavior compared with their accustomed human-human interaction. 
The findings of Reeves et al. suggest that human-machine interaction follows the
same rules as human-human interaction, cf. \cite{Reeves2003,
HandbookRobotics2016}. For this reason, task delegation must resemble the
communication observed in crafts enterprises which can often be reduced to a
triplet. Triplets are known among other
things from scene graphs where they are used to establish the relationship
between objects in an image via the following pattern: \texttt{<subject,
predicate, object>} \cite{Schroeder2019}.
Considering use case A, cf. \cref{sec:use-case_sanding}, the triplet
\texttt{<basin, sand, mineral cast>} could allow craftsmen to issue commands
without the need for programming.

The \gls{ui} must consider a noisy, dusty and temperature-variable environment
due to the fact that cobots are usually deployed in production facilities with
running tools creating dust, e.g. saw and sander, loud excavation systems and
the heat produced by machines.

Providing high-level task-based interaction capabilities
requires increased autonomy of the underlying system. In order to avoid
creating an unpredictable black box, the human operator should be provided with
up-to-date information and explanations about the current system status by the

\gls{ccu}. 
\subsection{Cognitive Control Unit}
\label{sec:ccu}

The \gls{ccu} is the central part of the software architecture and handles
several key tasks: (1) create and maintain a model of the environment, (2)
high-level task-based command decomposition, and (3) provide assistance for
human operators.

The \gls{ccu} must create and maintain an up-to-date model of the environment
of the machine as a single point of truth. Therefore, the \gls{ccu} must be
able to communicate with every other component in the system as well as
listen to the communication among other components to ensure correct
operation of other components. 

Additionally, the \gls{ccu} is responsible for decomposing a received
command from the human operator via the \gls{ui} into sub-tasks according to
provided build instructions created offline. The \gls{ccu} also contains
knowledge in the form of rules encoded offline. This knowledge is similar to
the knowledge apprentices gain during their education. The necessity is
provided by the abstract and high-level communication observed among craftsmen.

\Cref{fig:workingMemory} presents the Soar working memory structure required to
decompose a complex task into several sub-tasks utilizing \texttt{State},
\texttt{IO}, \texttt{Workspace}, \texttt{Object}, \texttt{Tool}, \texttt{Task}
and \texttt{Component} structures. The \texttt{State} and \texttt{IO} are  default Soar
structures and for this reason not described here.



The workspace serves as a top-level structure for the separation of physical
spaces where the robot could operate and allocate tools and objects to this
workspace. The workspace also contains links to all available tools and their
corresponding processes. In the future, restrictions with specific end-effectors 
could be enforced, e.g. restrict sanding-tool use in areas without
an excavation system. Applying this structure to the use case sanding
mineral-cast basins would result in a single workspace including a tool named
sander providing the production processes sand and polish. The object (basin)
allocated to the workspace has properties like name, material and might have
sub-elements of type \texttt{Task} and \texttt{BuildStructure}. 

Abstract high-level commands are represented in Soar as tasks with at least
three attributes: process, material, and object. Given a user-specified task
via the \gls{ui}, cf. \cref{sec:ui} the task is allocated to an object. 

\begin{figure}
    \centering
    \includegraphics[width=\linewidth, page=3]{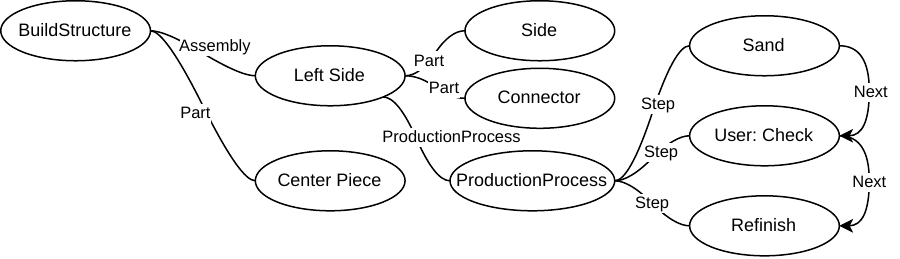}
    \caption{Reduced representation of the \texttt{BuildStructure} subtree with
    connected steps required to complete the build process. }
    \label{fig:buildstructure}
\end{figure}

In case task and object create a match, the \texttt{BuildStructure} is attached
to the \texttt{Object} and generates a linked tree of components based on
offline instructions. Each component can either be an assembly or a part, while
an assembly can consist of other assemblies and parts, similar to the structure
of a CAD model. Additionally, a \texttt{Component} has a
\texttt{ProductionProcess} node including \texttt{Step} nodes representing a
sub-task, which requires execution in order to complete a \texttt{Task}.
\Cref{fig:buildstructure} presents the \texttt{BuildStructure} node for the
use case sanding mineral cast basins. 

Besides the \texttt{Step} nodes resulting in sub-tasks being sent to the robot
for execution, sub-tasks can have different origins like pre- and
postcondition checks, cf. skill model by Schou et al. \cite{Schou2013}. 
One of the preconditions is to make sure that the correct
end effector is mounted on the robot. If the wrong end effector is mounted, the
\gls{ccu} sends a command to the relevant components to change the end
effector. The commands of the abstract high-level task result from the
generated \texttt{BuildStructure}, see \cref{fig:buildstructure}. Finally,
postconditions ensure that the task has been successfully completed, e.g.,
asking the operator to check the result. As a result, an abstract high-level
task is decomposed to multiple commands for the robot by the \gls{ccu}. 

Due to the symbolic representation of both knowledge and the current situation
in Soar, it is possible, that the \gls{ccu} explains its decisions and provides
information on the status of tasks currently being carried out. Thereby,
observed problems with highly automated systems like incomprehensible decisions
for human operators, cf. workload, unpredictability, and system competency
trade-off by Miller et al. \cite{Miller2000, Miller2007}, can be reduced.

\section{Contribution} 
\label{sec:contributions}
The project will have the following contributions to academia and industry:

\begin{itemize}
    \item test the hypothesis, that craftsmen prefer high-level task-based communication
        above mid-level skill-based programming and low-level instruction based
        programming when interacting with complex technical systems by means of
        a user study utilizing a demonstrator
    \item evaluate interaction methods and devices for the proposed triplet-based interaction concept 
        including touch-panel and virtual or augmented reality devices
    \item provide a description of use cases based on challenges observed in small crafts enterprises increasing 
        the variety of use cases compared with pick and place applications.
    \item technology transfer to \gls{sme} which lack research capabilities due to their small size
\end{itemize}

\section{Conclusion}


Observations in craft enterprises have revealed that craftsmen usually
communicate with each other and hand over tasks at an abstract and high level
that requires at least the knowledge of an apprentice. Based on this finding,
an assessment of the relevance of SMEs for the labor market, and a literature
review of relevant work, a concept for delegating tasks between humans and
machines was proposed. In contrast to current approaches on a skill-based
level, our approach aims at raising the level of abstraction and allow the
operator to specify complex tasks based on triplets in order to bridge the gap
between high-level task delegation and low-level robot instructions. The task
instructions are based on offline-generated knowledge of task execution. Based
on the use cases and the concept described in this paper, a demonstrator consisting of
the OnRobot Sander attached to an UR10e and an excavation system is
currently being built and implemented. 
As described above, it uses the cognitive architecture Soar to implement the interaction based on triplets.
The usability of the concept will be evaluated by
a user study utilizing the demonstrator.

\bibliographystyle{IEEEtran}
\bibliography{IEEEabrv,main.bib}

\end{document}